\documentclass[11pt]{article}

\usepackage[final]{acl}

\usepackage{times}
\usepackage{latexsym}

\usepackage[T1]{fontenc}

\usepackage[utf8]{inputenc}

\usepackage{microtype}

\usepackage{inconsolata}

\usepackage{graphicx}

\usepackage{tabularx}
\usepackage{xcolor}
\usepackage{tikz}
\usepackage{array}
\usepackage{multirow}

\usepackage{comment}
\definecolor{claim}{RGB}{17,122,101}
\title{Context-Aware Multimodal Claim Verification in Spoken Dialogues}

\author{
Chaewan Chun$^{1}$ \and
Delvin Ce Zhang$^{2}$ \and
Dongwon Lee$^{1}$ \\
$^{1}$The Pennsylvania State University, USA \\
$^{2}$University of Sheffield, UK \\
\texttt{czc5884@psu.edu, delvin.ce.zhang@sheffield.ac.uk, dongwon@psu.edu}
}

\begin{document}
\maketitle
\begin{abstract}
Spoken factual claims often occur within multi-turn conversations, where surrounding dialogue can provide context unavailable from the claim alone. Yet most fact-checking research evaluates isolated text, leaving conversational audio under-studied. We introduce MAD2, a synthetic Multi-turn Audio Dialogues benchmark for spoken claim verification with 1,000 two-speaker dialogues, 1,230 sentence-level check-worthy candidate annotations, and approximately 10 hours of audio. We also propose calibrated multimodal fusion of a context-aware audio encoder and a dialogue-aware text model. Adding dialogue context improves verification across settings, although the gains vary by scenario. Past-only context often approaches local offline performance, suggesting its usefulness when future context is unavailable. Fusion achieves its highest mean advantage over text with full-dialogue context, but does not consistently or significantly outperform text across settings. Under full-dialogue context, exploratory subgroup results show greater performance variation across dialogue scenarios for text and fusion, but across spread styles for audio.
\end{abstract}

\section{Introduction}
Millions hear factual claims in podcasts every day, yet most of those claims never enter the fact-checking pipeline. Audiences increasingly shift toward audio platforms for news and information~\cite{whittle}, yet most verification pipelines still assume written text~\cite{gear,kgat}. In text, claims can be linked to sources and cross-referenced via hyperlinks~\cite{transformer_xh,correct}; in podcasts and live streams, claims are embedded in multi-turn conversation without citations or attributions, and relevant context may be distributed across turns. Existing text benchmarks~\cite{fever, liar, dialfact} are not designed to capture this, leaving spoken, conversational misinformation underexplored.

What makes this problem particularly hard is that spoken misinformation is often \emph{constructed} through interaction. A statement that appears innocuous on its own may become misleading as speakers endorse, challenge, reinterpret, or elaborate on it. The surrounding exchanges may therefore provide verification cues unavailable to a claim-only model. Prior work on dialogue fact-checking has focused on adapting evidence-retrieval pipelines to better exploit textual dialogue context~\cite{dialfact,chamoun2023}, but has not examined whether the spoken signal itself carries veracity-relevant information. This leaves a basic empirical question open: does audio add information beyond the transcript, and which portions of the dialogue are most useful for verification?

\begin{figure}[t]
\centering
\begingroup
\footnotesize
\begin{minipage}{\columnwidth}
\raggedright
\textbf{Script: }
\textcolor{claim}{\textbf{Exactly. It's a real problem for the county budget. I mean, at some point, you have to ask if it's really ...}}

\textit{\textbf{ASR:}}
\textcolor{claim}{real problem for the county budget i mean at some point you have to ask if it s really ...}
\end{minipage}

\vspace{0.4em}

\begin{center}
\newcommand{\Tmax}{38.0}
\newcommand{\ClaimStart}{21.96}
\newcommand{\ClaimEnd}{29.73}
\pgfmathsetlengthmacro{\xunitMain}{0.85\linewidth/\Tmax}
\def\yunit{7mm}
\def\axisthick{1pt}
\def\claimline{1.4pt}

\begin{tikzpicture}[x=\xunitMain,y=\yunit,line cap=round]
  \draw[fill=black!10,draw=black!30,rounded corners=1pt]
        (0,0.1) rectangle (\Tmax,0.8);

  \draw[line width=\axisthick] (0,0.45) -- (\Tmax,0.45);

  \draw (0,0.62) -- (0,0.28) node[below=3pt] {0 s};
  \draw (\Tmax,0.62) -- (\Tmax,0.28)
        node[below=3pt] {\pgfmathprintnumber{\Tmax}~s};

  \draw[fill=claim,draw=claim,opacity=0.25,rounded corners=1pt]
        (\ClaimStart,0.1) rectangle (\ClaimEnd,0.8);
  \draw[claim,line width=\claimline]
        (\ClaimStart,0.45) -- (\ClaimEnd,0.45);

  \node[below=2pt,anchor=north,text=claim,font=\bfseries\footnotesize]
       at ({(\ClaimStart+\ClaimEnd)/2},0)
       {\pgfmathprintnumber{\ClaimStart}~s--\pgfmathprintnumber{\ClaimEnd}~s};
\end{tikzpicture}
\end{center}
\endgroup
\vspace{-0.7em}

\caption{Example of precise claim--audio alignment in \textbf{MAD2}, where a target claim is linked to its spoken segment using WhisperX~\cite{whisperx} word-level timestamps.}
\vspace{-0.2em}

\label{fig:mad2_claim_example}
\end{figure}
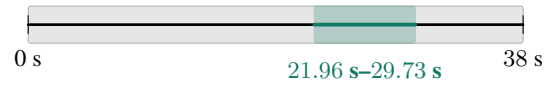

Progress on this question has been limited by two gaps. First, existing benchmarks do not combine conversational audio at scale with timestamped check-worthy candidates and veracity labels inherited from human fact-checked seed claims. Early efforts on political debates~\cite{kopev2019} and audio check-worthiness detection~\cite{ivanov2023} demonstrated that acoustic features can complement text for factuality prediction, but focused on monologue-style debate speech rather than conversational audio. Existing systems for live audio fact-checking~\cite{livefc} process broadcast-style streams through Automatic Speech Recognition (ASR) and text-based Natural Language Inference (NLI), without modeling acoustic features or dialogue context. Although text--image verification has a mature benchmark ecosystem~\cite{mocheg, mr2}, spoken conversational claims lack an equivalent resource. Second, systematic evidence remains limited on how dialogue context affects verification across modalities: whether it benefits audio and text equally, whether the modalities capture different aspects of the interaction, and how their performance changes as more of the conversation becomes available. These gaps leave the relationship among audio, transcripts, and dialogue context unresolved.


To address both gaps, we introduce \textbf{MAD2}\footnote{\url{https://huggingface.co/datasets/chaewanC/MAD2}}, a synthetic Multi-turn Audio Dialogues benchmark of 1{,}000 two-speaker dialogues (8{,}192 sentences; 1{,}230 sentence-level check-worthy candidate annotations) with approximately 10 hours of audio. Extending MAD~\cite{mad2025}, MAD2 adds two capabilities that the original benchmark lacked: higher-fidelity audio synthesis via MoonCast~\cite{mooncast} and precise claim-to-audio alignment via WhisperX~\cite{whisperx} word-level timestamps (Figure~\ref{fig:mad2_claim_example}). We also propose a calibrated conditional fusion of context-aware audio and text encoders.

Using MAD2, we systematically evaluate how conversational context affects verification across text-only, audio-only, and fusion models under matched sentence-based context windows. We focus on three questions: (i) how verification changes as context expands, (ii) whether past-only context can match local offline context, and (iii) when audio provides information beyond ASR transcripts. Our contributions are: (1) \textbf{MAD2}, a spoken-dialogue claim verification benchmark with paired audio, ASR transcripts, and claim-to-audio alignments, (2) a calibrated conditional fusion method, together with an analysis of when combining audio and text helps beyond text alone, and (3) a controlled evaluation protocol that systematically varies dialogue context under matched sentence-based windows across text-only, audio-only, and fusion models. The MAD2 dataset is publicly available on Hugging Face.


\section{Related Work}

Most fact-checking research operates over written claims, supported by benchmarks such as FEVER~\cite{fever}, LIAR~\cite{liar}, and DialFact~\cite{dialfact}, with dialogue work adapting evidence-retrieval pipelines to exploit textual context~\cite{dialfact,chamoun2023}. We inherit veracity labels from one such resource (LIAR) but depart from this line in modality: our task verifies claims from conversational \emph{audio}, where acoustic cues, disfluencies, timing, and prosody are available yet absent from transcript-only pipelines.

Work on spoken misinformation is comparatively limited. Studies on political debates and audio check-worthiness show that speech can carry useful signal~\cite{kopev2019,ivanov2023}, and recent systems verify spoken claims in streaming settings~\cite{livefc}. Unlike audio check-worthiness work, which decides whether a segment \emph{deserves} verification, our task predicts inherited seed-claim veracity in two-speaker dialogue and systematically tests how surrounding conversational context affects it; and unlike live systems that route verification entirely through ASR text, we model the acoustic signal directly. Large podcast corpora offer rich audio--text data~\cite{spotify,sporc} but lack timestamped check-worthy candidates paired with seed- or evidence-linked veracity labels, making them unsuitable for this evaluation. Finally, multimodal fact-checking has matured in text--image settings~\cite{mocheg,mr2}, but spoken dialogue remains underexplored. MAD2 fills this gap by pairing conversational audio with timestamped check-worthy candidates and inherited seed-claim veracity labels, enabling controlled evaluation of audio, text, and dialogue context in spoken claim verification.

\begin{table}[t]
\centering
\caption{Key extensions from MAD to MAD2.}
\label{tab:mad}
\footnotesize
\setlength{\tabcolsep}{3pt}
\begin{tabular}{p{2.2cm} p{2.4cm} p{2.2cm}}
\hline
\textbf{Dimension} & \textbf{MAD \newline \cite{mad2025}} & \textbf{MAD2} \\
\hline
Dataset size & 600 dialogues/ \newline 4.9k sentences & 1,000 dialogues/ 8.2k sentences \\
Check-worthy candidates & 1,748 total & 1,230 total \\
Inherited T/F labels & 810 T / 938 F & 537 T / 693 F \\
Audio model & XTTS-v2 & MoonCast \\
Audio quality & Flat, monotone & Natural,\newline conversational \\
Audio duration & Not reported & 10h 4m, avg.\ 36s \\
Transcript \newline alignment & Not available & WhisperX \\
\hline
\end{tabular}
\end{table}

\section{Dataset Creation: MAD2}

MAD2 is a synthetic English benchmark of two-speaker dialogues with sentence-level check-worthiness annotations, inherited seed-level veracity labels, and high-fidelity generated conversational audio. We adopt a controlled construction pipeline because obtaining expert veracity labels aligned to long-form podcast audio at scale is prohibitively expensive, and the synthetic design lets us vary how claims enter conversations and how speakers respond. We build on the construction pipeline of our prior work, MAD~\cite{mad2025}, and substantially extend it, as summarized in Table~\ref{tab:mad}. Compared with MAD, MAD2 combines broader dialogue coverage with more restrictive check-worthiness criteria, focusing on verifiable factual assertions that are relevant to public discourse and could cause harm if false.

Each dialogue is generated by conditioning Gemini 2.5 Pro~\cite{gemini} on a human fact-checked political claim from the LIAR~\cite{liar} benchmark. The dialogue generator receives the seed claim text but is not provided with its veracity label. The resulting dialogue receives the seed claim's true/false label. We sample 1,000 claims (500 true, 500 false) and generate one two-speaker script per claim, conditioned on fictional speaker profiles, a spread style, and a dialogue scenario. \textbf{Spread styles} capture how a claim enters conversation (\emph{Consequential Storytelling}, \emph{Casual Rumor-Mongering}, \emph{Ironic Dismissal}) and \textbf{dialogue scenarios} define the interactional dynamic (\emph{Collaborative Agreement}, \emph{Collaborative Skepticism}, \emph{The Classic Debate}, \emph{The Persuader and the Questioner}, \emph{The Unresolved Argument}). Each sentence is annotated with a binary check-worthiness label. These annotations identify which generated sentences express verifiable claim content; they do not introduce independent veracity judgments. Instead, every check-worthy candidate in a dialogue is paired with its single dialogue-level LIAR label. Automated quality control with ChatGPT~\cite{chatgpt} filters each dialogue for coherence and consistency with the seed claim. MAD2 comprises 1,000 dialogues (8,192 sentences; 1,230 sentence-level check-worthy candidate annotations: 537 true / 693 false) with about 10 hours of audio (avg.\ 36s/dialogue).

\textbf{Audio generation.} MAD used XTTS-v2\footnote{\url{https://huggingface.co/coqui/XTTS-v2}}, producing flat, monotone synthetic speech. MAD2 replaces it with MoonCast~\cite{mooncast}, a two-speaker podcast synthesis model designed to produce more natural prosody, conversational flow, and disfluencies. In its English podcast evaluation, MoonCast receives subjective scores of 4.54 for spontaneity and 4.50 for coherence, together with a word error rate of 1.81\%~\cite{mooncast}. MoonCast's reported WER uses a different evaluation corpus and ASR recognizer and is therefore not directly comparable with our MAD2 measurement. These results motivate our use of MoonCast as a closer synthetic approximation to podcast speech.

\textbf{Transcript alignment.} MAD2 provides WhisperX~\cite{whisperx} transcriptions with word-level timestamps. Since ASR output does not preserve sentence-level labels, we align each transcript back to the generated script to recover check-worthy annotations and timestamped claim spans. The alignment pipeline automatically flags low-scoring, partially matched, unmatched, and fallback or split matches; the saved alignments contain 90 flagged sentences across 77 dialogues. These ASR transcripts are used as text input in all experiments, matching realistic conditions where transcripts are automatically and imperfectly obtained. For an independent audio--script consistency check, we transcribe all 1,000 original recordings with Whisper large-v3~\cite{whisperv3} (faster-whisper, FP16, beam size 5), obtaining a normalized corpus-level WER of 4.21\% against the complete generation scripts. This score uses full-dialogue transcripts without script-conditioned sentence matching and applies identical normalization of case, punctuation, common contractions, and equivalent numeric expressions to both sides; it does not replace the WhisperX transcripts used in the experiments.

\textbf{Target-claim instances.}
While the script-generation protocol follows MAD~\cite{mad2025}, MAD2 turns each check-worthy script sentence into a timestamped target-claim instance. Each instance links the candidate sentence to its speaker, turn position, ASR transcript, dialogue-level veracity label, scenario type, spread style, and start--end time span in the waveform. This representation enables matched multimodal evaluation: the text model receives the ASR sentence window with the target claim marked, while the audio model receives the waveform crop spanning the same sentence window and retains the target's relative timestamp for claim-aware pooling. Thus, differences between text-only, audio-only, and fusion models reflect modality and context use rather than mismatched input scopes.

Modeling examples are defined only when a dialogue contains an eligible target-claim instance. A candidate enters the experimental pool if it is check-worthy, has an inherited binary seed label, and passes the alignment and ASR-quality gates applied to every model variant (alignment match score $\geq 0.8$ and average ASR confidence $\geq 0.8$). Dialogues without an eligible target remain part of the released benchmark but do not yield prediction instances. Of the 1,230 check-worthy candidates, all have binary inherited veracity labels; the alignment-score filter removes one candidate, and the subsequent ASR-confidence filter removes 98. The resulting experimental pool contains 1,131 target-claim instances from 845 dialogues.

\section{Methodology}

Given a dialogue and an aligned target-claim instance, the task is to predict the binary veracity label inherited from the dialogue's LIAR seed claim, $y \in \{0,1\}$ ($1$ = True, $0$ = False). We develop three model variants (audio-only, text-only, and fusion). The audio-only and text-only models are trained with cross-entropy loss, with checkpoints selected by validation AUC; fusion combines their predictions post-hoc. For each variant, the decision threshold is tuned on validation data by F1 score \cite{protoco} and applied to test data without further adjustment.

\subsection{Context Configurations}
\label{sec:sent-ctx}

We vary the scope and direction of dialogue context using sentence-window neighborhoods around each claim. Dialogues are segmented into ordered sentences by turn and position. For a claim at position $i$, we select a window of up to $N$ neighboring sentences on each side, preserving speaker turn structure. Here, $N$ indexes neighboring sentences rather than speaker turns, so even small windows such as $N=2$ can include adjacent cross-speaker context when claims occur near turn boundaries.
All three model variants operate over \emph{identical} sentence windows, enabling a fair cross-modal comparison.

\textbf{(1) Claim only} ($N=0$): the claim sentence alone, mimicking traditional fact-checking setups~\cite{kgat} where claims are verified in isolation.

\textbf{(2) Real-time} ($-N$): the target claim and up to $N$ preceding sentences, without future context, simulating live or streaming moderation. We evaluate $N \in \{1, 2\}$.

\textbf{(3) Offline} ($\pm N$): $N$ sentences on each side of the claim, reflecting post-hoc analysis of recorded audio. We evaluate $N \in \{1, 2\}$, with natural truncation at dialogue boundaries.

\textbf{(4) Full dialogue}: the entire dialogue containing the target claim, serving as the maximum-context offline condition.

\subsection{Audio-only Claim Verifier}
\label{sec:audio_model}

WavLM-base+ (94.68M parameters)~\cite{wavlmbase} is used as the speech encoder, supported by 
prior benchmarking~\cite{yang2024,ts-superb} and preliminary runs showing HuBERT~\cite{hubert}, 
wav2vec2~\cite{wav2vec2}, and Data2Vec~\cite{data2vec} yield weaker results. We extract a single 
contiguous audio crop spanning the sentence window using WhisperX word-level timestamps, with a 
$0.15$\,s padding margin on each side to avoid hard boundary effects, resampled to 16\,kHz and 
loaded on-the-fly during training.
The context crop and the claim span serve different roles. The waveform crop gives WavLM access to surrounding speech, pauses, speaker transitions, and local conversational context, while the claim timestamps identify which portion of that crop should drive the final decision. Thus, the model does not discard context before encoding, but it also does not ask the classifier to summarize the entire dialogue window uniformly. This mirrors the text model's claim boundary tokens: both modalities receive the same surrounding sentences, but both are given an explicit pointer to the target claim.

Rather than pooling uniformly over all frames, we apply \emph{claim-aware attention pooling} to focus the representation on the claim region while still letting the full context inform the encoder. We compute a binary mask over WavLM frame representations by mapping the claim's sample-level timestamps to frame indices via WavLM's convolutional downsampling ratio, then apply a learned attention pooler: a two-layer MLP with tanh activation \cite{prml} that produces a scalar score per frame, followed by softmax \cite{prml} over the masked frames to yield attention weights. The resulting pooled vector passes through LayerNorm and a two-layer classification head (Linear--ReLU--Dropout(0.1)--Linear)~\cite{bert} to produce binary logits.

\begin{figure}[t]
    \centering
    \includegraphics[width=\columnwidth]{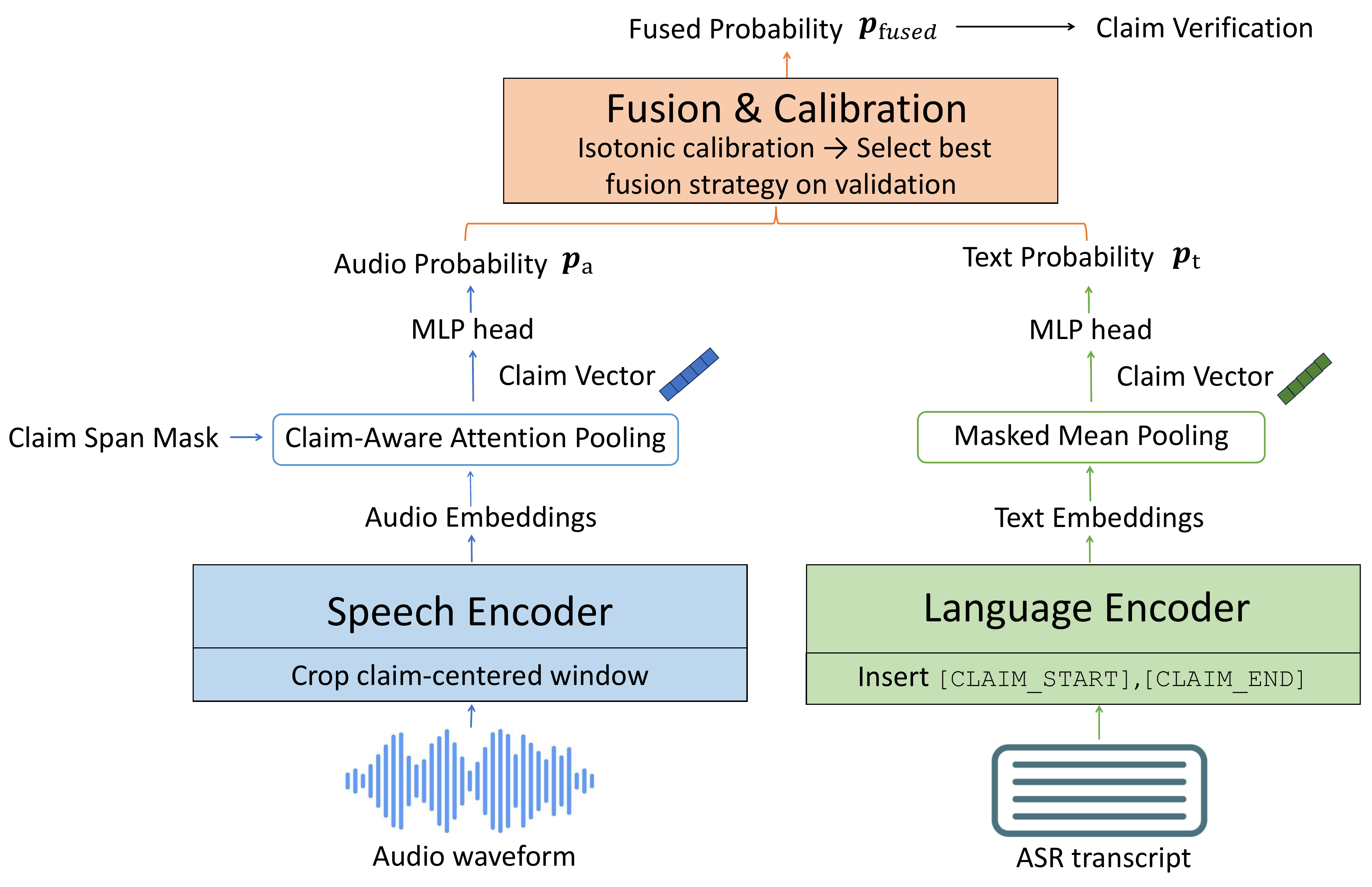}
    \caption{Architecture of the proposed multimodal claim verification system.}
    \label{fig:model}
\end{figure}

\subsection{Text-only Claim Verifier}
\label{sec:text_model}

The text-only model encodes the sentence window as a single serialized sequence using RoBERTa-base (125M parameters)~\cite{roberta}. To mark claim identity within the context, we wrap the claim sentence with special boundary tokens \texttt{[CLAIM\_START]} and \texttt{[CLAIM\_END]}, and prepend speaker tokens (\texttt{[SPK\_A]}, \texttt{[SPK\_B]}) and turn separators (\texttt{[TURN\_SEP]}) to preserve dialogue structure. All special tokens are added to the tokenizer vocabulary before fine-tuning. We obtain a sentence-window representation via masked mean pooling over the final-layer hidden states, then pass it through LayerNorm and the same two-layer MLP head as the audio model to produce binary logits.

\begin{table*}[t]
\centering
\caption{Claim verification results across context configurations and model variants, reported as mean $\pm$ standard deviation over three seeds.}
\label{tab:results_all}
\footnotesize
\setlength{\tabcolsep}{4pt}
\begin{tabular}{ll|cc|cc|cc}
\hline
\multicolumn{2}{c|}{\textbf{Setting}}
  & \multicolumn{2}{c|}{\textbf{Audio-only}}
  & \multicolumn{2}{c|}{\textbf{Text-only}}
  & \multicolumn{2}{c}{\textbf{Fusion}} \\
\cline{3-8}
\multicolumn{2}{c|}{}
  & F1 & AUC & F1 & AUC & F1 & AUC \\
\hline
\multicolumn{2}{c|}{Claim only}
  & 0.627 $\pm$ 0.015 & 0.651 $\pm$ 0.028
  & 0.687 $\pm$ 0.012 & 0.746 $\pm$ 0.011
  & 0.678 $\pm$ 0.009 & 0.727 $\pm$ 0.024 \\
\hline
\multirow{2}{*}{Real-time}
  & $-1$
  & 0.641 $\pm$ 0.018 & 0.682 $\pm$ 0.025
  & 0.649 $\pm$ 0.029 & 0.754 $\pm$ 0.037
  & 0.671 $\pm$ 0.025 & 0.735 $\pm$ 0.029 \\
  & $-2$
  & 0.658 $\pm$ 0.009 & 0.713 $\pm$ 0.016
  & 0.691 $\pm$ 0.043 & 0.779 $\pm$ 0.034
  & 0.699 $\pm$ 0.019 & 0.779 $\pm$ 0.040 \\
\hline
\multirow{2}{*}{Offline}
  & $\pm 1$
  & 0.643 $\pm$ 0.006 & 0.672 $\pm$ 0.015
  & 0.671 $\pm$ 0.020 & 0.767 $\pm$ 0.031
  & 0.689 $\pm$ 0.022 & 0.757 $\pm$ 0.040 \\
  & $\pm 2$
  & 0.633 $\pm$ 0.024 & 0.686 $\pm$ 0.022
  & 0.691 $\pm$ 0.028 & 0.780 $\pm$ 0.013
  & 0.692 $\pm$ 0.027 & 0.765 $\pm$ 0.011 \\
\hline
\multicolumn{2}{c|}{Full dialogue}
  & \textbf{0.707 $\pm$ 0.028} & \textbf{0.784 $\pm$ 0.018}
  & \textbf{0.721 $\pm$ 0.008} & \textbf{0.819 $\pm$ 0.032}
  & \textbf{0.735 $\pm$ 0.030} & \textbf{0.839 $\pm$ 0.020} \\
\hline
\end{tabular}
\end{table*}

\subsection{Calibrated Conditional Fusion}
\label{sec:fusion}

We use post-hoc fusion because preliminary joint-fusion variants, including a two-stage freeze-then-joint setup, did not improve on calibrated fusion in validation experiments. This may reflect the limited supervised scale of MAD2 for jointly fine-tuning two large encoders. Across context-aware settings, the independently trained modalities disagree on 20--38\% of test predictions, with audio correct in 33--53\% of those disagreements, calculated by aggregating counts across the three seeds within each setting. Disagreement is highest in the Real-time ($-2$) setting. We therefore evaluate selectively gated fusion strategies and calibrate both modalities onto a common probability scale before combining them. See Figure~\ref{fig:model}.

Audio may be useful in some settings without being uniformly beneficial. We therefore compare fixed-weight and conditional fusion strategies, selecting the fusion rule and its parameters separately for each context configuration and seed using validation data.

Let $p_t \in [0,1]$ denote the raw softmax probability of the True class output by the text model, and $p_a$ the corresponding probability from the audio model. Raw softmax probabilities are often poorly calibrated~\cite{calibration}:  text models tend toward overconfidence, audio toward underconfidence. We compare Platt scaling~\cite{platt}, isotonic regression, and no calibration, fitting calibrators separately for each modality using only the corresponding seed's validation predictions. Calibration and fusion settings are jointly selected by validation AUC; isotonic regression is selected in all context--seed combinations. Let $\hat{p}_t$ and $\hat{p}_a$ denote the probabilities after the selected calibration is applied.

Given $\hat{p}_t$ and $\hat{p}_a$, we search over five fusion strategies: (1) \emph{late fusion}: $p_f = \alpha \cdot \hat{p}_t + (1-\alpha) \cdot \hat{p}_a$, a fixed-weight average controlled by a grid-searched scalar $\alpha$; (2) \emph{recall boost}: $p_f = \hat{p}_t + \beta \cdot \max(0, \hat{p}_a - \hat{p}_t)$, which selectively raises the text probability when audio assigns a higher probability to the True class; (3) \emph{audio override}: replace $\hat{p}_t$ with a blend when $\hat{p}_t < \tau_t$ and $\hat{p}_a > \tau_a$; (4) \emph{conditional alpha}: $p_f = \alpha(\hat{p}_a) \cdot \hat{p}_t + (1 - \alpha(\hat{p}_a)) \cdot \hat{p}_a$, where $\alpha$ takes a lower value when $\hat{p}_a$ exceeds a probability threshold, increasing audio's weight dynamically; and (5) \emph{max-fusion}: $p_f = \max(\hat{p}_t, \gamma \hat{p}_a)$, with grid-searched $\gamma$. A text-only option ($p_f = \hat{p}_t$) is also included in the search. The best-performing strategy, calibration method, and hyperparameters are jointly selected by validation AUC.

For each seed, model hyperparameters, calibration, fusion strategy, and decision threshold are selected using only that seed's validation partition. Although dialogue assignments can overlap across seeds, no other seed's validation predictions are used for calibration or selection. The selected configuration is then evaluated on the corresponding seed's test partition. The selected fusion rules for each context and seed are reported in Appendix~\ref{app:fusion_rules}.

\section{Experiments}

\subsection{Implementation Details}
\label{sec:impl}

The audio-only and text-only models are fine-tuned with AdamW~\cite{adamw}. We search over eight hyperparameter combinations: learning rate $\in \{2{\times}10^{-5}, 3{\times}10^{-5}\}$, warmup ratio $\in \{0.05, 0.1\}$, and weight decay $\in \{0.01, 0.03\}$, with dropout fixed at 0.1 throughout. For each run, the best checkpoint is determined by validation AUC; the decision threshold is then tuned on the same validation predictions by F1 score. The hyperparameter configuration with the highest validation AUC is selected for final evaluation. The text model uses batch size 32 with bf16 mixed precision, while the audio model uses batch size 8 with gradient checkpointing to handle longer waveform crops. Both are trained for up to 30 epochs with early stopping (patience 5) on NVIDIA H100 GPUs. A full text model sweep (three configurations in parallel) takes approximately 1.5h wall-clock time, and audio model training takes approximately 5.7h per configuration.

Train, validation, and test partitions are dialogue-disjoint within each seed, although partition assignments overlap across seeds. Splits are stratified by a composite label of scenario type and spread style, so that each combination is proportionally represented across train, validation, and test; any combination with fewer than three dialogues is assigned entirely to train. Across seeds (42/43/44), train/val/test contain 588--591/127--128/126--130 dialogues (788--797/162--180/163--173 eligible target-claim instances), and results are averaged over three seeds with mean and standard deviation reported. The same quality filters and sentence-window definitions apply to all model variants. We assess AUC differences using 10{,}000 paired dialogue-level bootstrap resamples separately for each seed, with 95\% confidence intervals excluding zero indicating statistical significance. We report mean AUC and F1 and their standard deviations across seeds rather than computing these metrics from pooled predictions.

\subsection{Results and Analysis}
\label{sec:results}

We treat AUC as the primary reported metric, as it is threshold-independent, and report F1 for completeness. Table~\ref{tab:results_all} reports aggregate results 
across all context configurations and modalities, while Table~\ref{tab:stratified} breaks 
these down by scenario type and spread style to understand where context and modality matter 
most. Four findings emerge.

\textbf{(1) Context improves all modalities and narrows the text--audio gap.}
AUC improves from claim-only to full-dialogue context across all modalities: audio-only from $0.651$ to $0.784$, text-only from $0.746$ to $0.819$, and fusion from $0.727$ to $0.839$, yielding gains of $+0.133$, $+0.073$, and $+0.112$, respectively. Paired dialogue-level bootstrap intervals indicate significant improvements in all three seeds for audio and fusion, and in two of three seeds for text. Audio remains weaker than text, but benefits strongly from context: the text--audio gap narrows from $0.095$ at claim only to $0.035$ at full dialogue. These results suggest that surrounding conversational context is particularly useful for the audio model.

Context effects are broadly positive across scenario types in Table~\ref{tab:stratified}. Moving from claim-only to full-dialogue context increases mean AUC for all three modalities in all five scenarios, although the magnitude varies. The largest text-only gain occurs in Unresolved Argument, where AUC rises from $0.666$ to $0.858$. In Collaborative Skepticism, text-only performance remains nearly unchanged ($0.802$ to $0.804$), whereas audio and fusion improve from $0.675$ to $0.800$ and from $0.764$ to $0.883$, respectively. The effect of added context therefore differs across scenarios and modalities.

\textbf{(2) Real-time performance is competitive with offline.}
Models using only preceding sentences as context remain competitive with their offline counterparts. Audio-only at Real-time ($-2$) achieves AUC $= 0.713$ versus $0.686$ for Offline ($\pm 2$); text-only achieves $0.779$ versus $0.780$, and fusion achieves $0.779$ versus $0.765$. Paired dialogue-level bootstrap intervals include zero in all three seeds for each modality, providing no statistically significant evidence of a difference between these two settings. Audio-only also has higher mean AUC with past-only context at the smaller window size ($-1$: $0.682$ versus $\pm 1$: $0.672$). These results suggest that preceding discourse can provide much of the useful context available in the matched offline windows.

\textbf{(3) Fusion does not consistently outperform text.}
Fusion does not consistently improve AUC over text-only across context settings. It performs below text-only in four settings and is nearly tied with text-only under Real-time ($-2$), with both achieving AUC $= 0.779$. Its largest mean advantage occurs with full-dialogue context, where fusion reaches $0.839$ versus $0.819$ for text-only ($+0.020$). However, paired dialogue-level 95\% bootstrap confidence intervals for fusion-minus-text AUC include zero in every context--seed combination. Thus, the results suggest a possible full-dialogue benefit, but do not establish a statistically reliable improvement over text.

Exploratory stratified results show larger full-dialogue fusion advantages in selected categories. In Collaborative Skepticism, fusion achieves AUC $= 0.883$ versus $0.804$ for text-only ($+0.079$), the largest scenario-level advantage. Among spread styles, Casual Rumor-Mongering has the lowest text-only AUC ($0.773$) and the largest fusion advantage, reaching $0.818$ ($+0.045$). Fusion gains differ across dialogue categories, but these subgroup differences are descriptive rather than statistically established.

The gains are not uniform: fusion is nearly tied with text in Unresolved Argument ($0.861$ versus $0.858$) and performs below text in Classic Debate ($0.814$ versus $0.837$). At the aggregate full-dialogue level, fusion also has a higher mean F1 than text-only ($0.735$ versus $0.721$). These observations motivate further investigation of when multimodal fusion helps, rather than a general claim that audio consistently improves verification.

\textbf{(4) Text and fusion vary more by scenario; audio varies more by style.}
Under full-dialogue context, text-only and fusion performance vary more across dialogue scenarios than across spread styles, whereas audio-only performance shows the opposite pattern (Table~\ref{tab:stratified}). This suggests that the relative variation across the two category groupings depends on the modality. However, these descriptive comparisons do not establish whether scenario or style causally influences performance.

The two category groupings describe different aspects of a conversation: scenarios characterize how speakers interact, while spread styles characterize how a claim is introduced. Their differing patterns across modalities motivate reporting both breakdowns rather than relying only on aggregate performance. Whether particular interactional or acoustic cues explain these differences remains an open question.

\begin{table}[t]
\centering
\caption{Stratified AUC under claim-only (C) and full-dialogue (F) settings.}
\label{tab:stratified}
\footnotesize
\setlength{\tabcolsep}{3pt}
\begin{tabular}{@{}l|cc|cc|cc@{}}
\hline
\textbf{Category}
  & \multicolumn{2}{c|}{\textbf{Audio}}
  & \multicolumn{2}{c|}{\textbf{Text}}
  & \multicolumn{2}{c}{\textbf{Fusion}} \\
\cline{2-7}
 & C & F & C & F & C & F \\
\hline
\multicolumn{7}{@{}l}{\textit{Scenario Type}} \\
\hline
Collab. Agreement   & .663 & .751 & .773 & .792 & .766 & .814 \\
Collab. Skepticism  & .675 & .800 & .802 & .804 & .764 & .883 \\
Classic Debate      & .581 & .772 & .717 & .837 & .681 & .814 \\
Persuader \& Quest. & .645 & .808 & .773 & .780 & .738 & .852 \\
Unresolved Argument & .724 & .784 & .666 & .858 & .677 & .861 \\
\hline
\multicolumn{7}{@{}l}{\textit{Spread Style}} \\
\hline
Casual Rumor         & .624 & .795 & .669 & .773 & .668 & .818 \\
Conseq. Storytelling & .697 & .799 & .757 & .823 & .753 & .851 \\
Ironic Dismissal     & .602 & .730 & .778 & .827 & .708 & .834 \\
\hline
\end{tabular}
\end{table}


\section{Conclusion}

We presented MAD2 and a calibrated multimodal framework for spoken claim verification. Across modalities, conversational context improves verification, but the magnitude of that benefit depends on the dialogue scenario. Past-only context often approaches local offline performance, suggesting its usefulness for verification when future context is unavailable. Fusion shows its largest mean advantage over text with full-dialogue context, but the improvement is not statistically significant. Under full-dialogue context, exploratory subgroup results reveal different patterns by modality: text and fusion vary more across dialogue scenarios, while audio varies more across spread styles. Together, these findings show that systems for spoken claim verification should model conversational context selectively rather than assume that every additional turn or modality is uniformly useful.

\section*{Limitations}

MAD2 is synthetic and restricted to two-speaker English dialogues, so even with high-fidelity TTS it cannot fully capture the diversity, noise, speaker overlap, and interaction patterns of real podcast audio. Dialogue-generation patterns and synthesized prosody may correlate with inherited veracity labels, so performance on MAD2 may not transfer directly to naturally occurring conversations. In addition, although fusion shows advantages in some exploratory subgroups, this paper does not establish whether those differences arise from acoustic cues or other factors. Future work should extend the benchmark to longer, natural, and multilingual conversations, and should include deeper audio-centered analyses.

\section*{Ethical Considerations}

MAD2 is intended as a research benchmark for studying spoken claim verification, not as a deployable moderation system or an automated truth arbiter. The release documentation describes the dataset's provenance, generation methods, and intended research use. Models trained on synthetic audio may not generalize reliably to real speakers, accents, dialects, or noisy recording environments. Any real-world deployment would require human oversight, careful calibration, and evaluation on natural conversational audio.

\bibliography{custom}

\clearpage
\appendix
\section{Selected Fusion Rules}
\label{app:fusion_rules}

\begin{table}[ht]
\centering
\caption{Selected fusion rules for each context and seed. All configurations use isotonic calibration.}
\label{tab:selected_fusion_rules}
\footnotesize
\setlength{\tabcolsep}{3pt}
\begin{tabular}{@{}lccc@{}}
\hline
\textbf{Context} & \textbf{Seed 42} & \textbf{Seed 43} & \textbf{Seed 44} \\
\hline
Claim only & Late & Late & Conditional \\
Real-time ($-1$) & Conditional & Late & Conditional \\
Real-time ($-2$) & Conditional & Late & Conditional \\
Offline ($\pm 1$) & Recall boost & Late & Conditional \\
Offline ($\pm 2$) & Recall boost & Conditional & Conditional \\
Full dialogue & Conditional & Late & Late \\
\hline
\end{tabular}
\par\smallskip
\begin{minipage}{\columnwidth}
\footnotesize
Late denotes fixed-weight late fusion; Conditional denotes conditional-alpha fusion.
\end{minipage}
\end{table}

\end{document}